\documentclass[a4paper,fleqn]{cas-dc}
\usepackage[numbers]{natbib}
\usepackage{graphicx}  
\usepackage{amssymb}
\usepackage{amsfonts}
\def\tsc#1{\csdef{#1}{\textsc{\lowercase{#1}}\xspace}}
\tsc{WGM}
\tsc{QE}
\tsc{EP}
\tsc{PMS}
\tsc{BEC}
\tsc{DE}

\usepackage{CJKutf8}
\begin{document}
\begin{CJK}{UTF8}{gbsn}
\let\WriteBookmarks\relax
\def\floatpagepagefraction{1}
\def\textpagefraction{.001}



\title [mode = title]{\textbf{MSSC-BiMamba: Multimodal Sleep Stage Classification and Early Diagnosis of Sleep Disorders with Bidirectional Mamba}}




%
\author[1]{Chao Zhang}





\author[1]{Weirong Cui}
\ead{cuiweirong106@163.com}
\cormark[1]
\author[1]{Jingjing Guo}[orcid=0000-0002-4632-4364,]
\ead{jguo@mpu.edu.mo}
\cormark[1]
\affiliation[1]{organization={Centre in Artificial Intelligence Driven Drug Discovery, Faculty of Applied Sciences, Macao Polytechnic University},
     city={Macao},
     postcode={999078}, 
     country={China}}







\begin{abstract}
Monitoring sleep states is essential for evaluating sleep quality and diagnosing sleep disorders. Traditional manual staging is time-consuming and prone to subjective bias, often resulting in inconsistent outcomes. Here, we developed an automated model for sleep staging and disorder classification to enhance diagnostic accuracy and efficiency. Considering the characteristics of polysomnography (PSG) multi-lead sleep monitoring, we designed a multimodal sleep state classification model,  MSSC-BiMamba, that combines an Efficient  Channel Attention (ECA) mechanism with a Bidirectional State Space Model (BSSM). The ECA module allows for weighting data from different sensor channels, thereby amplifying the influence of diverse sensor inputs. Additionally, the implementation of bidirectional Mamba (BiMamba) enables the model to effectively capture the multidimensional features and long-range dependencies of PSG data. The developed model demonstrated impressive performance on sleep stage classification tasks on both the ISRUC-S3 and ISRUC-S1 datasets, respectively containing data with healthy and unhealthy sleep patterns. Also, the model exhibited a high accuracy for sleep health prediction when evaluated on a combined dataset consisting of ISRUC and Sleep-EDF. Our model, which can effectively handle diverse sleep conditions, is the first to apply BiMamba to sleep staging with multimodal PSG data, showing substantial gains in computational and memory efficiency over traditional Transformer-style models. This method enhances sleep health management by making monitoring more accessible and extending advanced healthcare through innovative technology.

\end{abstract}



\begin{keywords}
sleep stage classification \sep 
Mamba\sep
early diagnosis of sleep disorders \sep
efficient channel attention \sep
bidirectional state space model
\end{keywords}

\maketitle

\section{Introduction}

Sleep, a fundamental aspect of human health, involves multiple critical stages in cognitive function and overall well-being \cite{peter2015sleep}. The accurate classification of sleep stages and assessment of sleep health are vital for diagnosing and treating sleep disorders. Traditionally, polysomnography (PSG) has been the gold standard for monitoring sleep stages; however, it requires specialized equipment and is generally confined to clinic settings. This limitation underscores the need for more accessible and less intrusive methods of sleep analysis. 

Recent advances in machine learning, particularly deep learning, have paved the way for innovative approaches to sleep stage classification  \cite{loh2020automated}. Among these, the transformer model \cite{vaswani2017attention}, known for its effectiveness in handling sequential data primarily in the field of natural language processing, offers promising avenues for sleep data analysis due to its ability to capture long-range dependencies in time-series data. However, the transformer's self-attention mechanism, which evaluates interactions across all sequence positions, significantly slows down as data length increases, and hence suffers from low efficiency due to heavy computational demands and limited scalability with large datasets. This necessitates more efficient algorithms or adaptations of existing models to optimize performance in tasks like sleep stage classification and anomaly detection without compromising computational resources. 

Modern state space models (SSMs) particularly effective in capturing long-range dependencies, have evolved significantly with recent innovations. The Mamba model \cite{gu2023mamba} advances SSMs by integrating time-varying parameters and a hardware-aware algorithm, enhancing training and inference efficiency dramatically. Despite its superior scalability and potential as an alternative to Transformers in language modeling, Mamba is limited by its unidirectional approach and lack of positional awareness. 

Addressing these limitations, the Vision Mamba \cite{zhu2024vision} employs the Bidirectional State Space Model (BSSM) to leverage dynamics in both forward and reverse directions, significantly boosting prediction accuracy for sequential tasks. This bidirectional enhancement not only ensures high accuracy but also enhances efficiency, making Vision Mamba ideal for large-scale studies and potentially suitable for real-time sleep monitoring applications. 

Our current work builds on these advancements. By integrating the strengths of both the Transformer and the Vision Mamba's BSSM, our approach effectively captures complex temporal patterns in sleep data, facilitating precise sleep stage classification and sleep health discrimination. The bidirectional capability of the BSSM, combined with the deep learning prowess of the Transformer, allows for a robust analysis of large-scale, multimodal sleep monitoring datasets.

The implications of this work are profound for both clinical practice and personal health monitoring. By enhancing sleep stage classification and health assessment through an accessible and efficient model, our methodology bridges the gap between intricate clinical evaluations and practical, real-time monitoring solutions. This transition not only improves the accessibility of health monitoring but also ensures that advanced healthcare interventions are more attainable for the general population, promoting better sleep health management through innovative technology.

The main contributions of this paper are as follows:

\begin{itemize}

\item \textbf{Innovative Model Architecture}: We introduce the MSSC-BiMamba model with an innovative architecture, which combines the Efficient Channel Attention mechanism with BiMamba, tailored specifically for sleep stage classification and health status discrimination.

\item \textbf{Enhanced Efficiency}: The MSSC-BiMamba model, employing complex multimodal PSG data, significantly improves computational and memory efficiency, and bridges the gap between intricate clinical evaluations and practical, real-time monitoring solutions. 

\item \textbf{Superior Results}: Our model demonstrates superior performance on the ISRUC-S1 and ISRUC-S3 datasets for sleep stage classification and achieves a high accuracy of 0.952 on the Sleep-EDF153 and ISRUC datasets for sleep health detection, thereby confirming its generalizability and effectiveness across different sleep-related tasks.

\end{itemize}  


\section{ Related work}
\subsection{Classification of sleep stages}

Traditionally, the classification of sleep stages is performed by experienced sleep specialists or physicians, who categorize 30-second or specific time intervals of polysomnographic (PSG) data into various sleep stages according to established sleep assessment criteria \cite{liang2012rule}. This process is known to be both laborious and time-consuming \cite{malhotra2013performance}. In contrast, machine learning algorithms may require significantly shorter durations to accomplish the same classification task \cite{chang2019ultra}, while experts have traditionally extracted features from the time domain \cite{sharma2017automatic}, frequency domain \cite{zoubek2007feature}, and time-frequency domain \cite{al2019detection} to preprocess data for machine learning methods. The selection of effective features remains a critical issue in enhancing the classification performance of traditional classifiers.

The emergence of deep learning has further advanced the automation process, enabling direct extraction of complex features from raw data, thus reducing the preprocessing workload \cite{sekkal2022automatic}. Therefore, an increasing number of deep learning methods are being applied to sleep stage classification tasks, including CNNs \cite{phan2018joint}, RNNs \cite{zhu2020ocrnn}, GCNs \cite{jia2021multi}, and others. However, CNNs may struggle to effectively capture long-term dependencies in time-series data. While RNNs are capable of handling sequential data and capturing long-term dependencies in time series, they are prone to issues such as vanishing or exploding gradients during training. Additionally, GCNs exhibit lower efficiency in processing large-scale graph data.

Ji et al. \cite{10233098, JI2024107992} proposed several sleep stage classification models. These studies transformed polysomnographic (PSG) data into the frequency domain space, combined with time-domain signals as input to the model. They then utilized 3D CNNs, 2D CNNs, and GCN networks for classification, achieving satisfactory results. In comparison to these methods, we propose an approach that solely utilizes time-domain signals but achieves superior performance.

 \subsection{State space model and Mamba}
 
Recent research advances have led to a surge of interest in state-space models (SSMs). Originating from the classical Kalman filtering model \cite{kalman1960new}, SSMs excel at capturing long-term dependencies. Researchers have proposed several SSM-based methods, such as Structured State Space Sequence Models (S4) \cite{gu2021efficiently} and S4D \cite{smith2022simplified}, for handling sequential data from various tasks and modalities, particularly in modeling remote dependencies. Due to their convolutional and near-linear computations, they exhibit high efficiency in processing long sequences. In recent studies, Mamba \cite{gu2023mamba} integrated time-varying parameters into SSMs and proposed a hardware-aware algorithm to achieve highly efficient training and inference. Compared to the original Mamba, Bidirectional Mamba offers higher efficiency and performance. Based on this structure, we explore the temporal PSG signals.

\section{Methods}
\subsection{Overview of the framework}
\begin{figure*}[ht]
	\centering
		\includegraphics[width=1\linewidth]{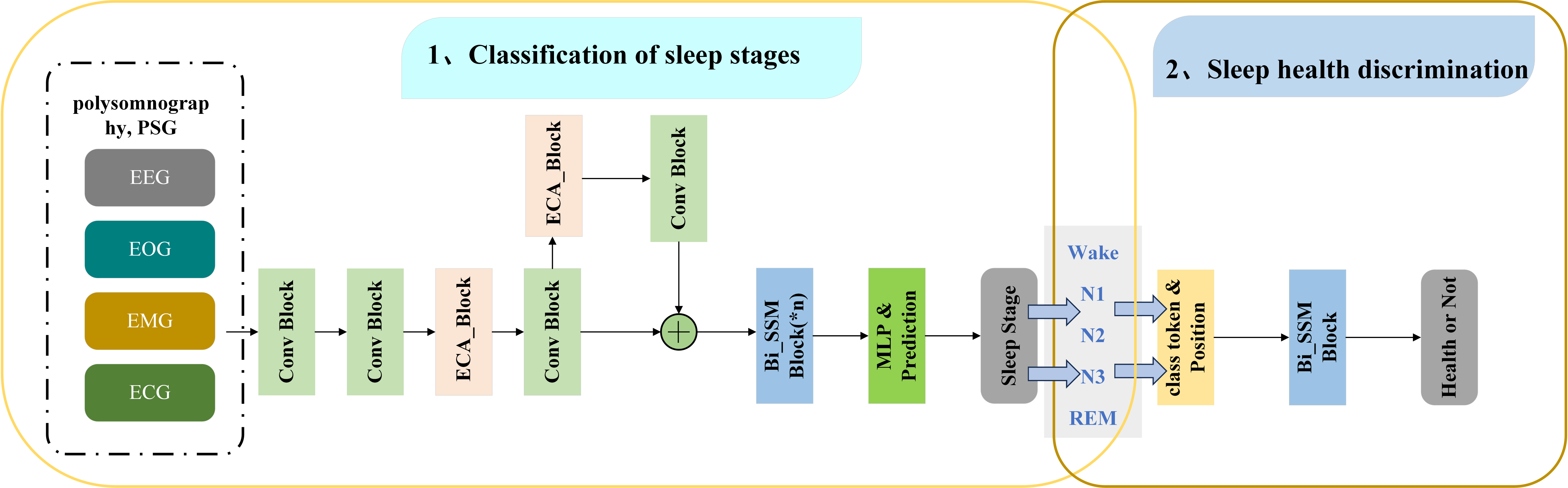}
	\caption{The architecture of the proposed MSSC-BiMamba model and its two tasks: 1) sleep stage classification using PSG data; 2) leveraging sleep stage information to identify sleep disorders.}
\label{fig:model}

\end{figure*}

The proposed MSSC-BiMamba model, depicted in Figure \ref{fig:model}, tackles two pivotal tasks within the realm of sleep medicine: 1) sleep stage classification using PSG data, such as EEG and EOG signals; 2) leveraging overnight sleep stage data to determine whether the participant is in a healthy sleep state. This innovative model is specifically designed to enhance diagnostic accuracy and efficiency, thereby boosting a deeper understanding of sleep health in clinical environments, and facilitating the early diagnosis of sleep disorders.

The architecture of MSSC-BiMamba incorporates the Efficient Channel Attention (ECA) module to focus on salient features in the data effectively, followed by the Bidirectional Mamba ({BiMamba}) module, which processes these features to capture complex temporal relationships. This setup ensures a robust analysis of sleep patterns. In the experimental setup, we detail the configuration of hyperparameters and the specific conditions under which the model operates. We conclude by describing the evaluation metrics used to assess the performance of the model, ensuring a comprehensive understanding of its effectiveness in clinical applications.

\subsection{The Efficient Channel Attention module}

In recent years, the introduction of channel attention mechanisms into convolutional blocks has garnered widespread attention. A prominent method in this domain is SENet (Squeeze-and-Excitation Networks)\cite{hu2018squeeze}, which learns channel attention for each convolutional block and significantly enhances the performance of various deep CNN architectures. While these methods achieve high accuracy, they often lead to increased model complexity and substantial computational burden. To address these challenges, Wang et al.\cite{wang2020eca} proposed the Efficient Channel Attention (ECA) module. The ECA module avoids dimensionality reduction and effectively captures cross-channel interactions, ensuring both efficiency and effectiveness. This module has achieved commendable results in the field of image processing.

\begin{figure}[ht]
	\centering
		\includegraphics[width=1\linewidth]{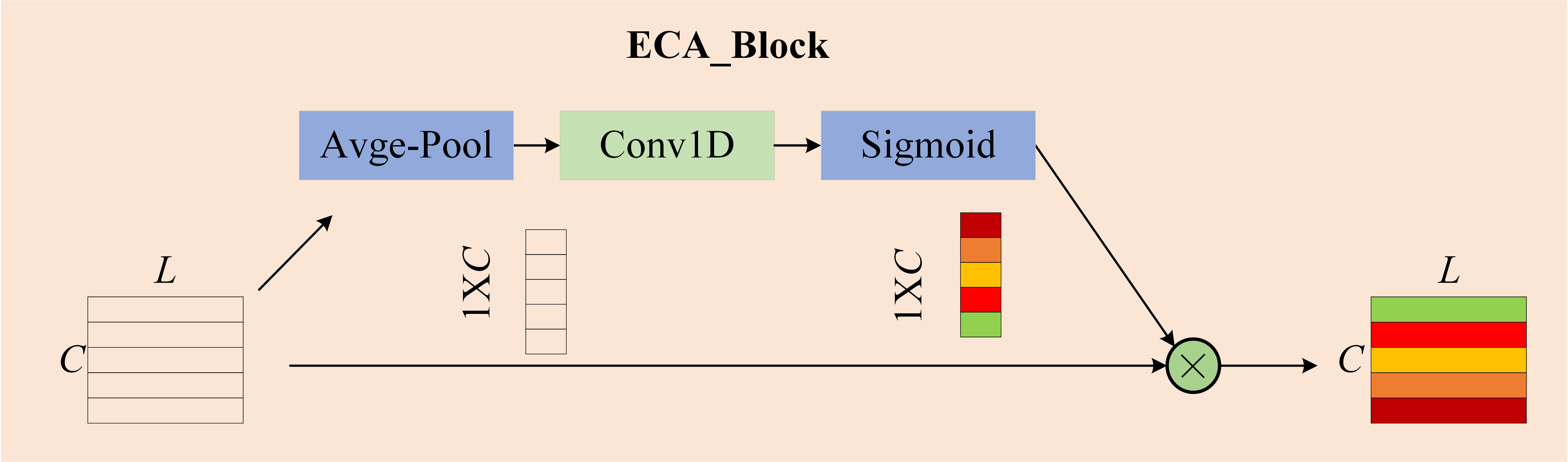}
	\caption{The structure of the ECA module.}
\label{fig:ECA}
\end{figure}

The channel attention mechanism is critical for dynamically adjusting the channel responses of feature maps by learning the importance of each channel. This process enhances the representational capacity of neural networks by allowing for the dynamic recalibration of feature channel correlations. Specifically, for time series data, this involves adjusting multiple PSG channel data to capture temporal patterns and dependencies more effectively. Building on the ECA module, we propose an adaptation tailored to time series data (Figure \ref{fig:ECA}). Our approach ensures the suitability and efficacy of the ECA mechanism in handling the unique characteristics of time series, such as temporal dependencies and sequential patterns.

Let ${x_{ci}}$ denote the input feature map, where \textit{c} is the number of channels and \textit{L} represents the length of the feature map. First, the global spatial information of each channel is computed through a global average pooling (GAP) operation, resulting in channel descriptors ${S_c}$.

\begin{align}{S}_{c}=\frac{1}{L}\sum\limits_{i=1}^{L}{{x}_{ci}}\end{align}

Subsequently, channel weights $W$ are obtained through convolutional layers and activation functions. 

\begin{align}
    W=\sigma (Conv1d(S))
\end{align}

Finally, the learned channel weights are applied to the original feature map, yielding the weighted feature map ${\tilde{X}_{ci}}$.
\begin{align}
    {{\tilde{X}}_{ci}}={{W}_{c}}\cdot {{X}_{ci}}
\end{align}

In this way, the network can strengthen the important channel features for sleep stage classification tasks while suppressing those unimportant channel features, thereby enhancing the model's ability to classify sleep stages.
\subsection{Bidirectional Mamba}

The S4 and Mamba structures discretize the state space representation of continuous systems. They utilize zero-order hold (ZOH) to maintain the dynamic characteristics of the system, addressing the limitation of direct implementation of continuous systems on digital computers, as digital computers can only process discrete signals. Through precise sampling and holding processes, stable and efficient digital implementation of continuous systems is permitted. The continuous state \textit{h}(\textit{t})  and input  \textit{x}(\textit{t}) can be represented as: 

\begin{align}
    h'(t) &= Ah(t) + Bx(t) \\
    y(t) &= Ch(t) + Dx(t) 
\end{align}


where \textit{A} is the state transition matrix,  \textit{B} is the input matrix, \textit{C} is the output matrix and \textit{D} is the direct transmission matrix. 


Through the time scale parameter $\Delta$, the continuous parameters \textit{A} and \textit{B} are transformed into discrete parameters, resulting in $\bar{A}$, $\bar{B}$. Zero-order hold (ZOH) is a method of maintaining the value of a signal unchanged during sampling and holding until the next sampling point. After discretizing  \textit{A} and \textit{B}, the discrete versions using the time scale parameter  $\Delta$ can be rewritten in the following form:

\begin{align}
\begin{matrix}
     {{h}_{t}}  =\bar{A}{{h}_{t-1}}+\bar{B}{{x}_{t}} \\
{{y}_{t}}  =C{{h}_{t}}
\end{matrix}
\end{align}

where, $\bar{A}  =\exp (\text{ }\!\!\Delta\!\!\text{ }A)$, $\bar{B}  ={{(\text{ }\!\!\Delta\!\!\text{ }A)}^{-1}}(\exp (\text{ }\!\!\Delta\!\!\text{ }A)-I)\cdot B$, finally, the model's output is obtained through a global convolution.

\begin{align}
\begin{matrix}
   \bar{K}=(C\bar{B},C\overline{AB},\cdots ,C{{{\bar{A}}^{M-1}}}\bar{B}), \\ 
  y=x\cdot \bar{K} \\ 
 \end{matrix}
\end{align}
where $\bar{K}$ is the structured convolutional kernel, and ${M}$ is the length of the input sequence ${x}$.

\begin{figure}[ht]
	\centering
		\includegraphics[width=1\linewidth]{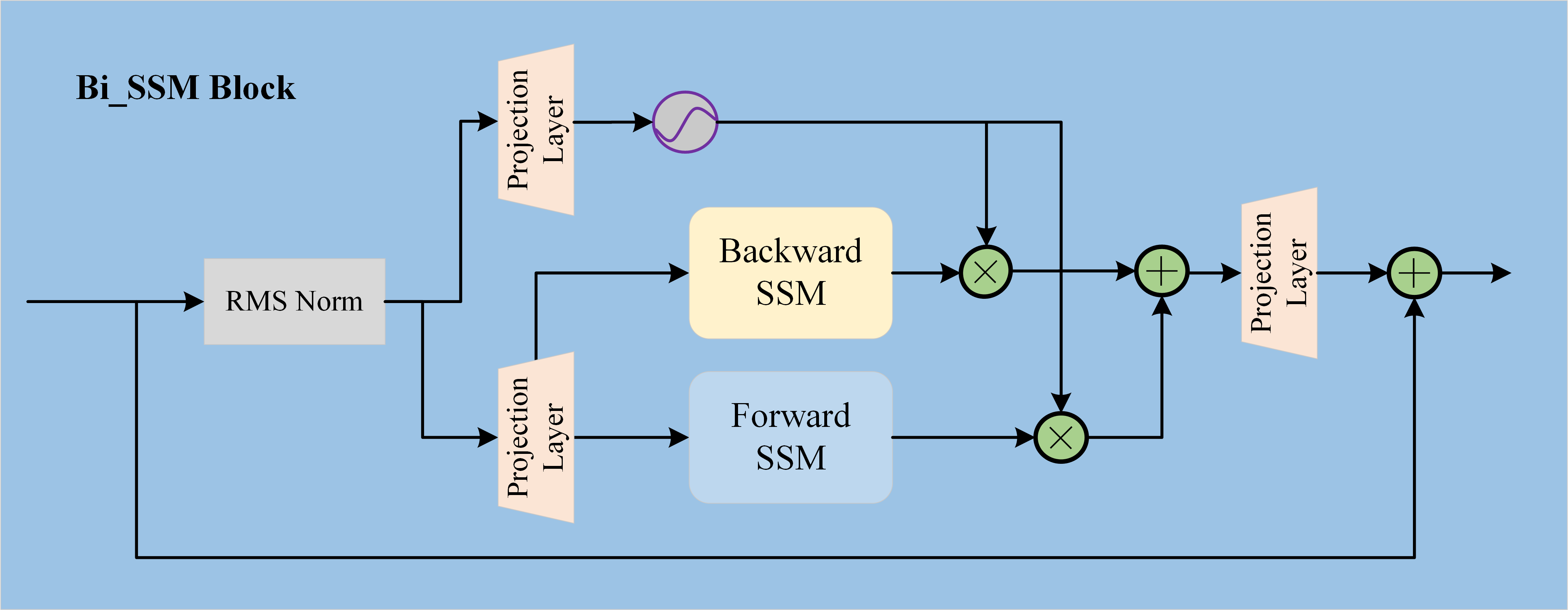}
	\caption{The structure of the BSSM Block.}
 \label{fig:bi_ssm}
\end{figure}

As shown in Figure \ref{fig:bi_ssm}, the bidirectional Mamba utilizes both forward and backward modules to expedite the SSM process and enhance the capability of acquiring contextual information. Specifically, it scans the sequence once from start to end and again from end to start. Subsequently, it averages the outputs from both scans to obtain a comprehensive representation.

\subsection{Evaluation metrics}


Accuracy measures the percentage of correctly predicted results relative to the total sample size.

\begin{align}
    Accuracy=\frac{TP+TN}{TP+FN+FP+TN}
\end{align}
Precision calculates the proportion of samples correctly predicted as positive class out of all samples predicted as positive class.
\begin{align}
    precision=\frac{TP}{TP+FP}
\end{align}
Recall is calculated as the proportion of samples that were correctly predicted as positive by the model out of all actual positive samples, using the function below:
\begin{align}
    recall=\frac{TP}{TP+FN}
\end{align}

in which, TP stands for true positive, TN stands for true negative, FP stands for false positive, and FN stands for false negative.

The F1 score is the harmonic mean of precision and recall, utilized to strike a balance between the two.

\begin{align}
    F1=\frac{2\times recall\times precision}{recall+precision}
\end{align}

The Kappa coefficient (Kappa) is an index that considers random agreement, used to measure the accuracy of a classifier, computed using the following equation:
\begin{align}
    \kappa =\frac{{{p}_{o}}-{{p}_{e}}}{1-{{p}_{e}}}
\end{align}

in which, ${p}_{o}$ represents the overall accuracy of the model. ${p}_{e}$ denotes the probability of chance agreement.

\begin{table}[width=.9\linewidth,cols=3,pos=h]
\caption{Dataset information. }\label{tbl:Health-Cls}
\begin{tabular*}{\tblwidth}{@{}ccc@{}}
\toprule
Subset&	Sleep Health	&Subject Number\\
\midrule
ISRUC-S1	&Unheadlthy	&100\\
ISRUC-S3	&Headlthy&	10\\
Sleep-EDF153	&Headlthy	&153\\
\bottomrule
\end{tabular*}
\end{table}

\subsection{Datasets}
The ISRUC-Sleep \cite{khalighi2016isruc} and Sleep-EDF \cite{867928} datasets summarized in Table \ref{tbl:Health-Cls} are used in the present work. The ISRUC-Sleep dataset \cite{khalighi2016isruc} comprises data from both healthy subjects and individuals with sleep disorders. Specifically, the ISRUC-S3 subset contains data from 10 healthy subjects, while the S1 subset includes data from 100 participants with sleep disorders. The data provider performed preprocessing on all polysomnography (PSG) recordings, which involved the following steps: 1) 50 Hz power-line noise removal using a notch filter; 2) for electroencephalogram (EEG) and electrooculogram (EOG) data, a bandpass Butterworth filter was applied to obtain waves in the frequency range of 0.3 Hz to 35 Hz; 3) electromyogram (EMG) data were filtered using a low cutoff frequency of 10 Hz and a high cutoff frequency of 70 Hz. Due to the presence of noise, the last 30 epochs from each subject were removed. These preprocessing steps ensured the quality and reliability of the dataset for subsequent analyses and sleep stage classification.

\begin{table*}[ht]
\caption{Detailed information of the ISRUC-Sleep datasets used for sleep stage classification.}\label{tbl:IS-datasets}
\begin{tabular*}{\textwidth}{@{\extracolsep{\fill}}ccccccccc@{}}
\toprule
Subset&Subject Number&W&N1&N2&N3&REM&Total&K-fold \\
\midrule
ISRUC-S3&10&1674&1217&2616&2016&1066&8589&10\\
ISRUC-S1 (50)&50&10097&5555&13250&8675&5779&43356&25\\
ISRUC-S1 (all)&100&20098&11062&27511&17251&11265&87187&-\\
\bottomrule
\end{tabular*}
\end{table*}

The Sleep-EDF dataset \cite{867928} contains 197 whole-night PolySomnoGraphic sleep recordings. The 153 SC* files (SC = Sleep Cassette) were obtained in a 1987-1991 study of age effects on sleep in healthy Caucasians aged 25-101, without any sleep-related medication.

\subsection{Experimental setup}

According to the standards set by the American Academy of Sleep Medicine (AASM)\cite{berry2012aasm}, sleep can be divided into five different stages. These stages include wakefulness (W), rapid eye movement (REM) sleep, and three non-rapid eye movement (NREM) stages, namely N1, N2, and N3.

For the classification of sleep stages, we utilized all data from the S3 dataset and 50 participants with odd-numbered IDs from the S1 dataset (Table \ref{tbl:IS-datasets}), and selected PSG data from ten channels, including EEG, EOG, and ECG (Table  \ref{tbl:IS-Channel}).

\begin{table}[width=.9\linewidth,cols=2,pos=h]
\caption{Channel data selection explanation.}\label{tbl:IS-Channel}
\begin{tabular*}{\tblwidth}{@{} LL@{} }
\toprule
Channel &Name\\
\midrule
EOG & LOC-A2,ROC-A1\\
EEG & 1F3-A2,C3-A2,O1-A2, F4-A1,C4-A1,O2-A1\\
Chin EEG & X1 \\
ECG& X2\\
\midrule
SUM& 10\\
\bottomrule
\end{tabular*}
\end{table}
We initially downsampled from 200 Hz to 100 Hz, followed by slicing operations on the data of each channel. The data input into the network comprised 1000 points (30 s), about 33.33 Hz. The ISRUC-S3 dataset employed a 10-fold cross-validation, while the ISRUC-S1 dataset utilized a 25-fold cross-validation. Cross-validation was independently performed for each subject to ensure that data from the same subject did not appear simultaneously in both the training and validation sets. The experimental setup and hyperparameters are summarized in Table \ref{tbl:experimental}. 

\begin{table}[width=.9\linewidth,cols=2,pos=h]
\caption{Experimental environment and hyperparameter descriptions.}\label{tbl:experimental}
\begin{tabular*}{\tblwidth}{@{} LL@{} }
\toprule
Parameter	& Value\\
\midrule
GPU	&NVIDIA GeForce RTX 4090\\
CPU	&AMD Ryzen 9 7900 12-Core Processor\\
Pytorch	&Torch 2.1.1+cu118\\
Python&	3.10.13\\
Epoch&	40\\
Batch Size	&100\\
Learning Rate	&0.001\\
Weight Decay&	0.0001\\
Dropout&	0.2\\
Optimizer&	Adam\\

\bottomrule
\end{tabular*}
\end{table}

For the sleep health prediction, all datasets in Table \ref{tbl:Health-Cls} are used. The original sleep labels were subjected to one-hot encoding, and then the data for the first maximum length of 850 sleep cycles (850*30s, approximately 7.08 hours) were taken. To test the performance of different training and validation set ratios, we experimented with five ratios. By changing the different data ratios, we ensured that the proportion of each part of the data, that is, the proportion of health and sleep disorders, was the same during the division.

\section{Experiments and numerical analysis}

\subsection{Classification experiments on the ISRUC-S3 and ISRUC-S1 datasets}

To assess the performance of our model, we conducted experiments using four different combinations on the ISRUC-S3 dataset, testing the functionality of different modules. The experimental results (Table \ref{tbl:S3compare}) indicate that incorporating the ECA module can notably enhance performance. Compared to the model without the ECA module, performance improvements are observed across all metrics except for F1 in the W stage, where the performance remains relatively consistent. To evaluate the performance of models with different depths, we configured bimamba modules with 1 layer, 2 layers, 3 layers, and 10 layers. Through comparison, we found that increasing the depth of the model does not necessarily lead to performance improvement. However, for the challenging N1 stage with a limited sample size, some improvement is observed. The model with CNN, ECA and 1-layer bimamba performed best. 

When compared to previous models (Table \ref{tbl:S3compare}), our model outperforms them in most performance metrics on the ISRUC-S3 dataset. However, it exhibits marginally lower F1-scores in the Wake (W) and Rapid Eye Movement (REM) stages compared to the JK-STGCN \cite{ji2022jumping} and MixSleepNet \cite{JI2024107992} models. Additionally, our model benefits from a significantly reduced parameter count, making it more efficient than earlier deep-learning models in terms of computational resources and memory usage. This efficiency makes our model particularly suitable for deployment in clinical settings where computational power may be limited.

\begin{table*}[ht]
\caption{Comparison with other methods on the ISRUC-S3 dataset.}\label{tbl:S3compare}
\begin{tabular*}{\textwidth}{@{\extracolsep{\fill}}cccccccccc@{}}
\toprule
\multirow{2}{*}{Model}& \multirow{2}{*}{Parameter}& \multicolumn{3}{c}{Overall Metrics} & \multicolumn{5}{c}{Per-class F1-score (F1)}\\
\cmidrule(lr){3-5} \cmidrule(lr){6-10}
& & ACC & F1 & Kappa & W & N1 & N2 & N3 & REM \\

\midrule
RF \cite{memar2017novel}&<0.1M& 0.702 & 0.685 & 0.616 & 0.838 & 0.470 & 0.671 & 0.763 & 0.684  \\
     DeepSleepNet \cite{supratak2017deepsleepnet}&21M & 0.719 & 0.696 & 0.643 & 0.831 & 0.463 & 0.742 & 0.851 & 0.595  \\  
GraphSleepNet \cite{jia2020graphsleepnet} &-& 0.786 & 0.770 & 0.724 & 0.864 & 0.540 & 0.782 & 0.869 & 0.793 \\
  JK-STGCN \cite{ji2022jumping} &- & 0.831 & 0.814 & 0.782 & \textbf{0.900} & 0.598 & 0.826 & 0.901 & 0.845 \\
 MixSleepNet \cite{JI2024107992}&2.4M & 0.830 & 0.821 & 0.782 & 0.899 & 0.625 & 0.819 & 0.899 & \textbf{0.860 } \\
\midrule
CNN+1bimamba & 0.47M&0.845  & 0.817  & 0.794  & 0.871  & 0.620  & 0.834  & 0.910  & 0.852   \\ 
        CNN+ECA+1bimamba& \textbf{0.47M}& \textbf{0.852} & \textbf{0.824} &\textbf{ 0.803} & 0.886 & 0.624 & \textbf{0.841} & 0.915 & 0.854  \\ 
        CNN+ECA+2bimamba&0.73M & 0.847  & 0.819  & 0.796  & 0.878  & 0.625  & 0.831  & 0.914  & 0.847   \\ 
        CNN+ECA+3bimamba&0.99M & 0.850  & 0.822  & 0.800  & 0.871 & 0.629  & 0.840  & 0.915  & 0.855  \\ 
        CNN+ECA+10bimamba &2.79M& 0.850  & 0.823  & 0.800  & 0.879  & \textbf{0.635 } & 0.838  & \textbf{0.916}  & 0.846 \\ 
\bottomrule
\end{tabular*}
\end{table*}

\begin{figure}[ht]
	\centering
		\includegraphics[width=1\linewidth]{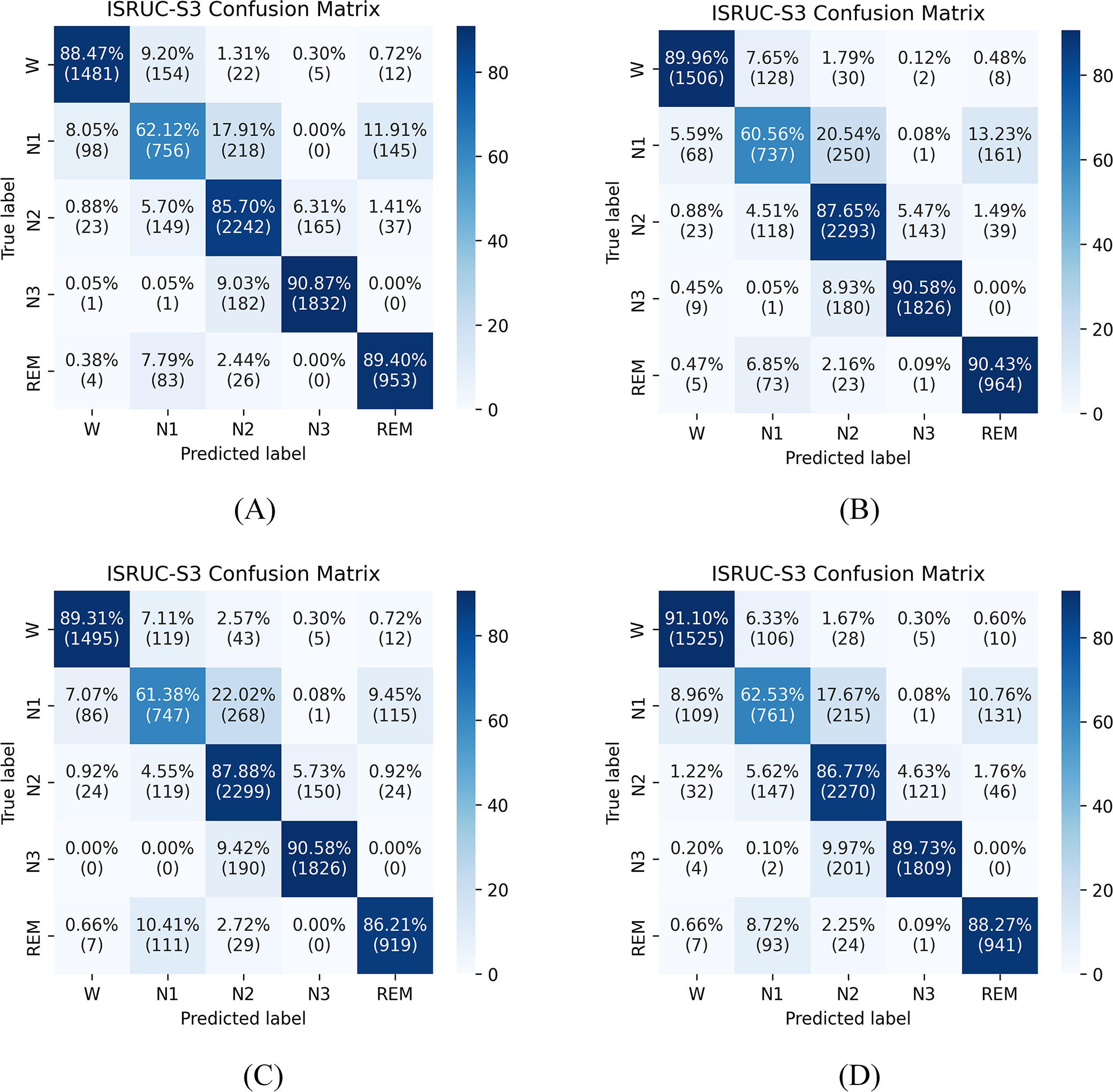}
	\caption{The confusion matrices for experiments on the ISRUC-S3 dataset: A, CNN+1bimamba; B, CNN+ECA+1bimamba; C, CNN+ECA+2bimamba; D, CNN+ECA+10bimamba.}
 \label{Matrix}
\end{figure}

The confusion matrices in Figure \ref{Matrix} illustrate the performance of different models on the ISRUC-S3 dataset. By comparing panels (A) and (B), it is evident that the inclusion of the Efficient Channel Attention (ECA) module enhances classification performance for all stages except N1. Further comparison of panels (B), (C), and (D) reveals that deepening the network improves the recognition performance for the N1 stage. However, this improvement comes at the cost of reduced performance for other stages.

After conducting a comparative analysis on the ISRUC-S3 dataset, we finalized the model parameters and selected the CNN+ECA+1bimamba model. Subsequently, we performed experiments on the ISRUC-S1 dataset. As illustrated in Table \ref{tbl1:compareS1}, our model outperforms other models in most performance metrics. Although it exhibits slightly lower F1-scores in the W and N1 stages compared to Mixsleepnet and JK-STGCN, it excels in other stages, highlighting its overall effectiveness. Given that the model parameters were optimized for ISRUC-S3, it is likely that the performance on ISRUC-S1 could be further enhanced with additional tuning specific to this dataset. Nonetheless, the current results underscore the potential and adaptability of the model in diverse sleep stage classification tasks.

\begin{table*}[!h]
\caption{Comparison with Other Methods on the ISRUC-S1 Dataset.}
\label{tbl1:compareS1}
\begin{tabular*}{\textwidth}{@{\extracolsep{\fill}}ccccccccc@{}}
\toprule
\multirow{2}{*}{Model} & \multicolumn{3}{c}{Overall Metrics} & \multicolumn{5}{c}{Per-class F1-score (F1)}\\
\cmidrule(lr){2-4} \cmidrule(lr){5-9}
 & ACC & F1 & Kappa & W & N1 & N2 & N3 & REM \\
 
\midrule
 RF \cite{memar2017novel}& 0.699  & 0.649  & 0.607  & 0.841  & 0.307  & 0.705  & 0.750  & 0.640   \\
DeepSleepNet\cite{supratak2017deepsleepnet} & 0.730  & 0.691  & 0.654  & 0.850  & 0.385  & 0.739  & 0.830  & 0.648   \\
GraphSleepNet\cite{jia2020graphsleepnet} & 0.780  & 0.751  & 0.715  & 0.889  & 0.463  & 0.763  & 0.825  & 0.813   \\    
JK-STGCN\cite{ji2022jumping} & 0.820  & 0.798  & 0.767  & 0.895  & \textbf{0.550}  & 0.811  & 0.883  & 0.850   \\ 
Mixsleepnet\cite{JI2024107992} & 0.813  & 0.787  & 0.757  & \textbf{0.908 } & 0.512  & 0.799  & 0.871  & 0.844   \\
\midrule
CNN+ECA+1bimamba & \textbf{0.830}  & \textbf{0.801}  & \textbf{0.773}  & 0.901  & 0.547  & \textbf{0.812}  & \textbf{0.885}  &\textbf{ 0.860}   \\ 
\bottomrule
\end{tabular*}
\end{table*}

\begin{figure*}[ht]
	\centering
		\includegraphics[width=\linewidth]{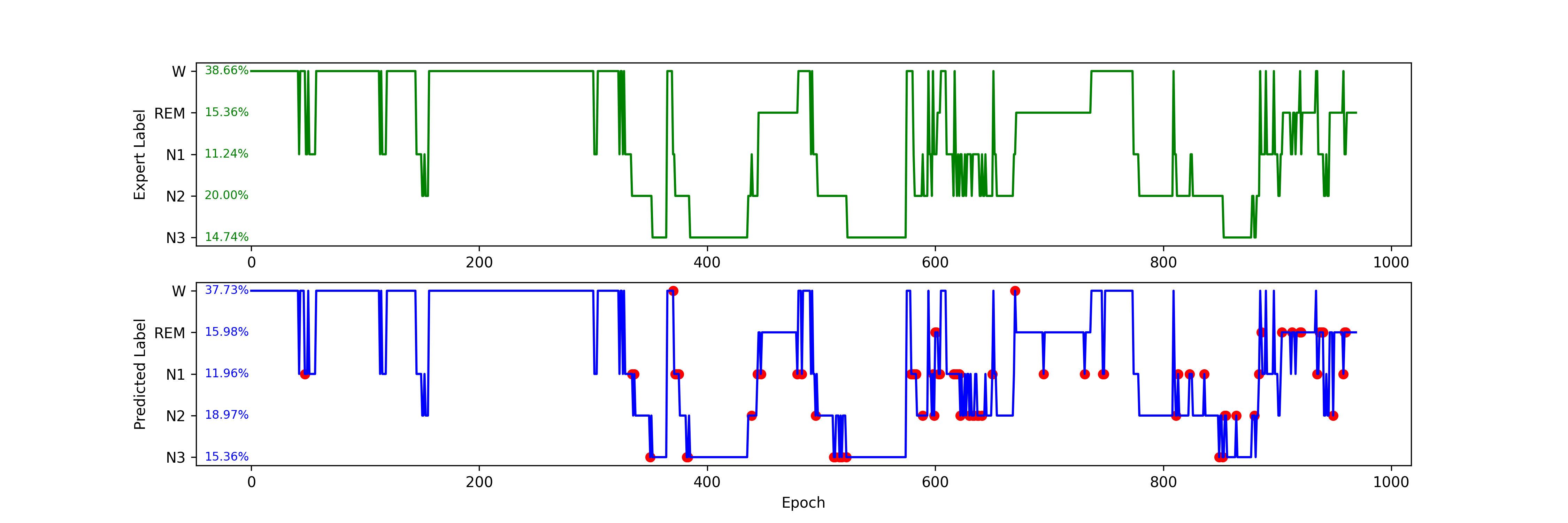}
	\caption{Visualization comparison of sleep stage classifications between expert-labeled outcomes and predictions by our S3 model for participant 8 in the ISRUC-S3 dataset.}
 \label{fig:s3-8}
\end{figure*}
\begin{figure*}[ht]
	\centering
		\includegraphics[width=\linewidth]{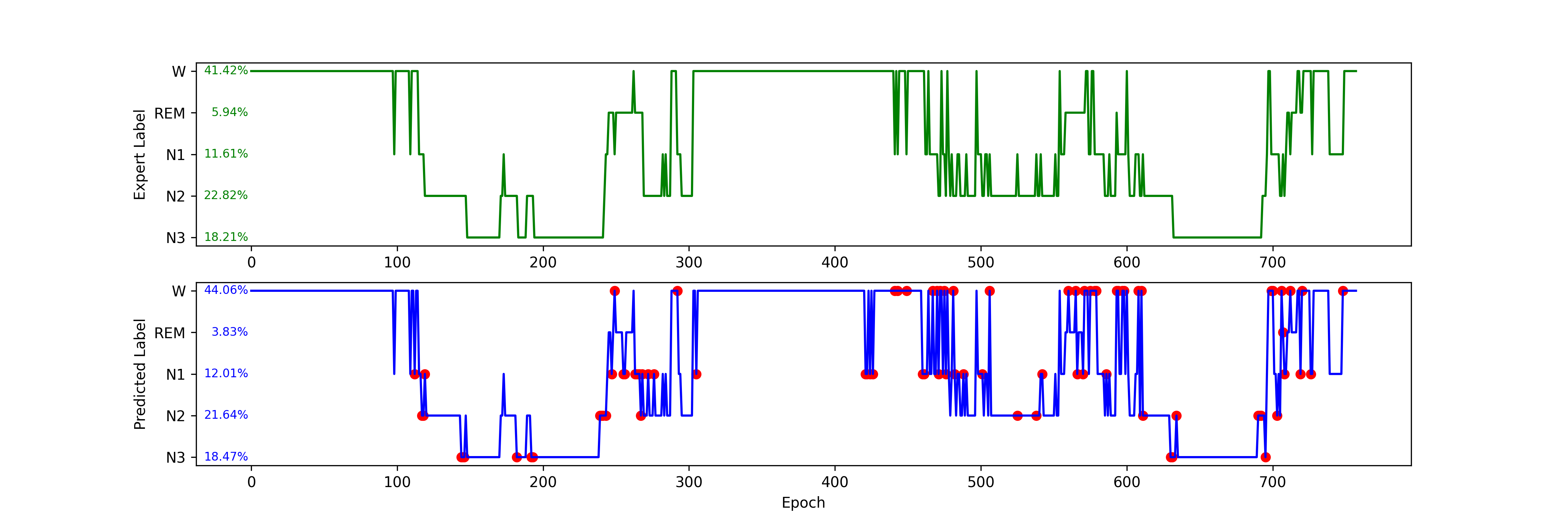}
	\caption{Visualization comparison of sleep stage classifications between expert-labeled outcomes and predictions by our S1 model for participant 39 in the ISRUC-S1 dataset.}
 \label{fig:s1-39}
\end{figure*}
\begin{figure*}[ht]
	\centering
		\includegraphics[width=\linewidth]{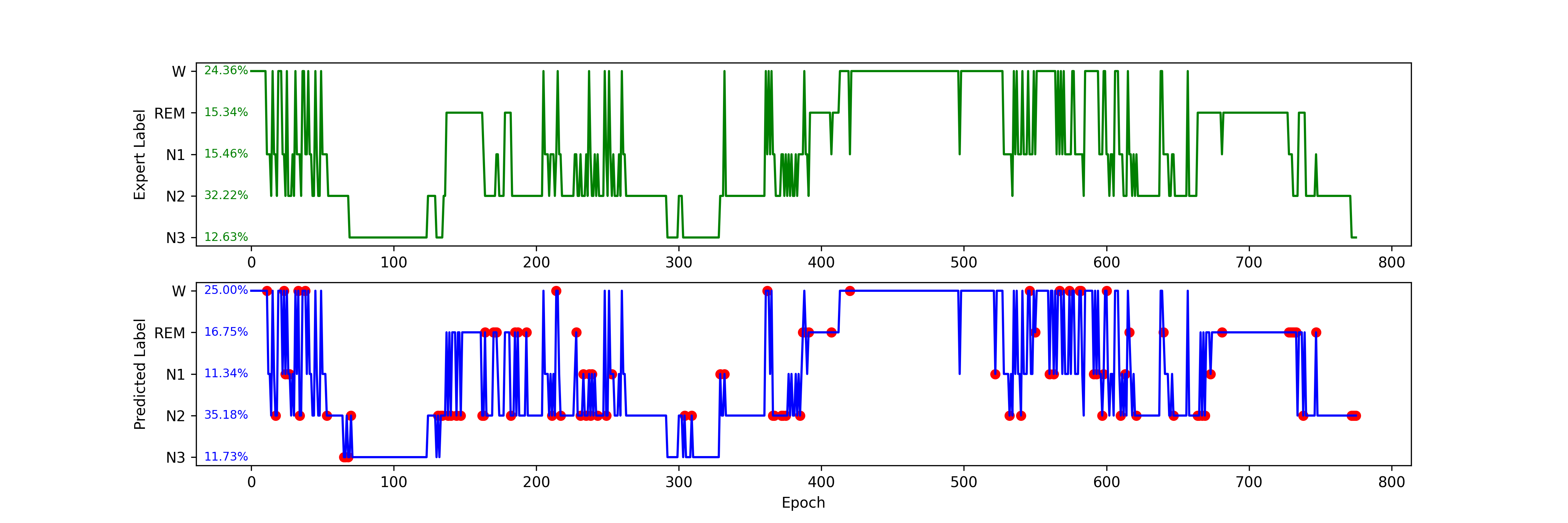}
	\caption{Visualization comparison of sleep stage classifications between expert-labeled outcomes and predictions by our S1 model for participant 41 in the ISRUC-S1 dataset.}
 \label{fig:s1-41}
\end{figure*}

Figures \ref{fig:s3-8}-\ref{fig:s1-41} illustrate the sleep state classifications for participants, comparing expert-labeled outcomes with those produced by our models. While there are some discrepancies during transitions between sleep stages, the majority of the classification results align closely with expert assessments.

Therefore, the developed model demonstrated impressive performance on sleep stage classification tasks on both the ISRUC-S3 and ISRUC-S1 datasets. These datasets respectively contain clinical PSG data from populations with healthy and unhealthy sleep patterns. This distinction highlights the model's capability to effectively differentiate and analyze sleep stages across diverse health conditions, showcasing its potential for broad applications in sleep medicine. Especially, our efficient model can free clinical physicians from the tedious task of manual staging, allowing them to focus more on supervising and correcting the automated classifications, as well as devoting more energy to the treatment and care of patients. This automated classification method not only enhances diagnostic accuracy, but also significantly improves efficiency when dealing with large volumes of data, bringing significant technological innovations to the field of sleep medicine.


\subsection{Cross-validation experiments}

In order to assess the generalization ability of the model across different datasets, we performed cross-validation experiments using models trained on ISRUC-S1(50) and ISRUC-S3 datasets, utilizing S1(50), S1(100), and the entire S3 dataset as test sets to evaluate the performance of all models through K-fold validation, and finally computed the average performance.

\begin{table*}[ht]
\caption{Comparison of cross-experiment results between the ISRUC-S3 and ISRUC-S1 models.}\label{tbl:s1s3}
\begin{tabular*}{\textwidth}{@{\extracolsep{\fill}}cccccccccc@{}}
\toprule
\multirow{2}{*}{Model} & \multirow{2}{*}{Datasets} &\multicolumn{3}{c}{Overall Metrics} & \multicolumn{5}{c}{Per-class F1-score (F1)}\\
\cmidrule(lr){3-5} \cmidrule(lr){6-10}
 & & ACC & F1 & Kappa & W & N1 & N2 & N3 & REM \\
 
\midrule
\multirow{2}{*}{S3} &S3&0.857&0.851&0.817&0.931&0.706&0.840&0.893&0.885\\
& S1(50)  &0.330 &0.281 &0.160&0.389 &0.148 &0.225 &0.368 &	0.274\\
\midrule
\multirow{3}{*}{S1(50)}&S1(50) &0.852 &	0.833&0.809&0.931 &	0.603&0.836&0.902&0.892\\
&S1(100) &0.808	 &0.788 &0.752 &0.899&0.534&0.794&0.862&0.849\\
&S3 &0.797&0.774&0.737&0.890&0.516&0.794&0.876&0.796\\
        \bottomrule
\end{tabular*}
\end{table*}

From Table \ref{tbl:s1s3}, it can be observed that the S1(50) model performs well on all three subsets, S1(50), S1(100), and S3, demonstrating its strong generalization ability. However, models trained on the S3 dataset do not perform as well on the S1 dataset, possibly due to differences between the S3 and S1 datasets, as well as the smaller size of the S3 dataset. Overall, the model's robust performance across various metrics underscores its potential for reliable and accurate sleep stage classification. 

\begin{figure*}[ht]
	\centering
		\includegraphics[width=1\linewidth]{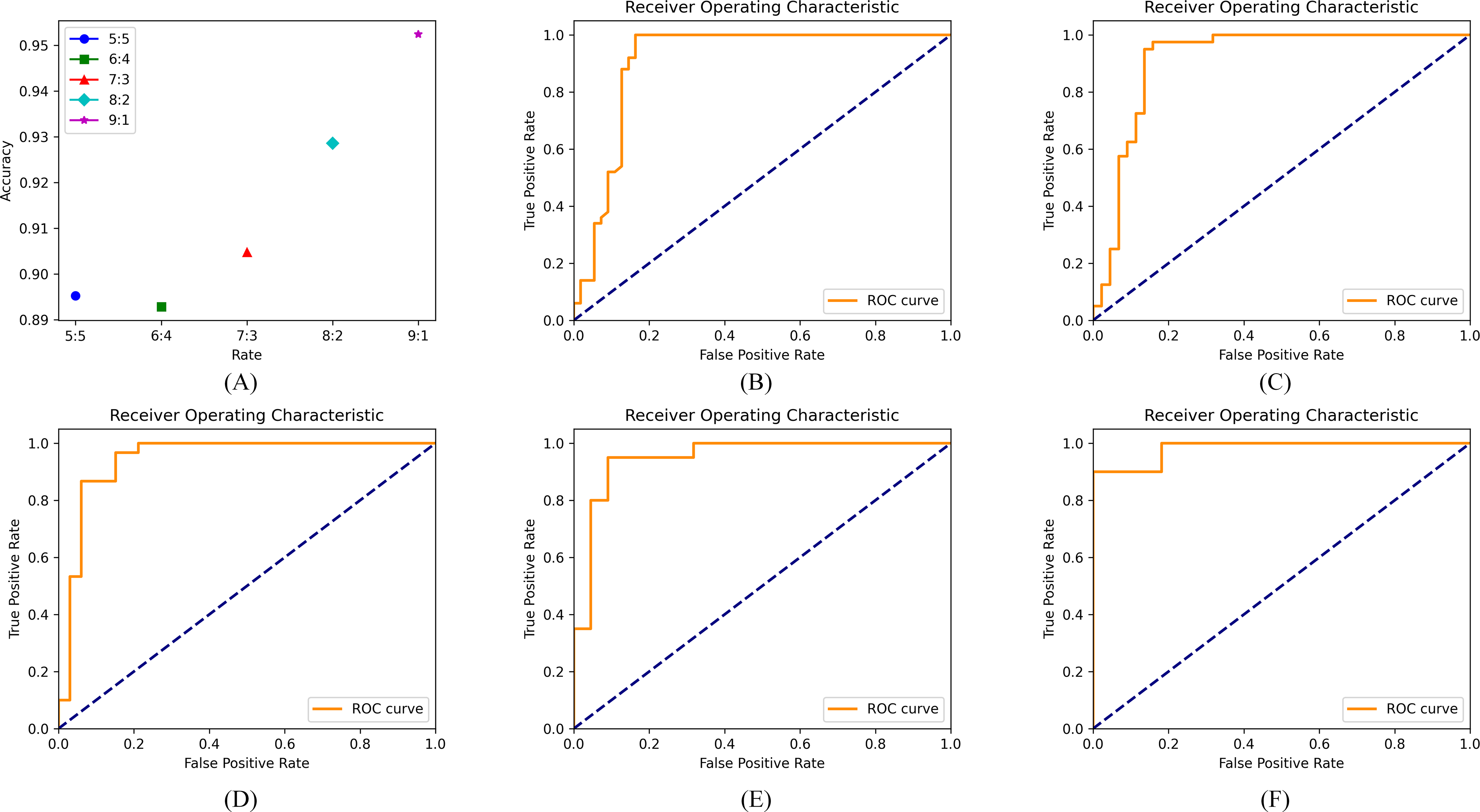}
	\caption{Comparison of sleep health prediction models: A, Accuracy of various models as the proportion of the training set increases; B-F, ROC curves for these models at increasing the training set sizes.}
 \label{acc_ROC}
\end{figure*}

\subsection{ Sleep health discrimination}

We conducted sleep health status prediction using data from the first one hundred participants in the Sleep-EDF153 dataset, as well as the ISRUC-S1 and ISRUC-S3 datasets. In total, the dataset comprised 110 instances of healthy sleep and 100 instances of sleep disorders. We validated different training-testing ratios to assess the prediction performance. As shown in Figure \ref{acc_ROC} of experimental results, increasing the proportion of the training set leads to improved performance of the model, demonstrating its robustness and accuracy. When the training set and test set were split 9:1, the model achieved an accuracy of 0.952. This high level of accuracy underscores the potential of our model to significantly aid in the early detection and classification of sleep health issues, thereby facilitating timely and effective interventions in clinical settings.

\section{Conclusion}

The deep learning framework powered by Mamba has demonstrated impressive generalization capabilities when applied to data related to sleep disorders. This characteristic allows the model to not only adapt efficiently to the training data but also maintain excellent predictive performance on previously unseen data. Such generalization ability is crucial for the model's reliability and stability in real-world applications. This paper preliminarily validates the effectiveness of the Mamba-driven deep learning framework model in predicting sleep health disorders, showcasing its potential in identifying and predicting various types of sleep disorders. However, sleep disorders are numerous, and each type exhibits different characteristics and manifestations. To further enhance the model's prediction accuracy, we plan to conduct more detailed analyses and training on different types of sleep disorders in future research. By incorporating more enriched features and optimizing algorithms, we believe that the Mamba-driven deep learning framework model can achieve more precise and timely predictions for various sleep disorders, thereby providing stronger support in the field of sleep health. In summary, the Mamba-powered sleep stage recognition model exhibits not only robust generalization capabilities but also impressive timeliness in handling complex and diverse sleep disorder data. This advancement holds promise for providing new solutions to assist doctors in annotation, effectively transforming medical experts from front-end operators to back-end supervisors. Future research will further explore its potential, advancing progress in sleep medicine.

\section*{Declaration of competing interest}

The authors declare that they have no known competing financial interests or personal relationships that could have appeared to influence the work reported in this paper.

\section*{Data availability}

The data that support the findings of this study are available from the corresponding author upon reasonable request.

\section*{Acknowledgments}
This research is funded by Macao Polytechnic University (\text{RP/CAI-01/2023}).

\printcredits

\bibliographystyle{cas-model2-names}

\bibliography{cas-dc}




\end{CJK}
\end{document}